\pdfoutput=1

\documentclass[11pt]{article}
\usepackage{CJKutf8}

\usepackage[preprint]{acl}

\usepackage{times}
\usepackage{latexsym}

\usepackage[T1]{fontenc}

\usepackage[utf8]{inputenc}
\usepackage{subfig}

\usepackage{microtype}

\usepackage{inconsolata}

\usepackage{graphicx}

\usepackage{csquotes}

\title{Limpeh ga li gong: Challenges in Singlish Annotations}

\author{Luo Qi Chan\textsuperscript{*} \\
  Carnegie Mellon University \\
  \texttt{luoqic@andrew.cmu.edu} \\\And
  Lynnette Hui Xian Ng\textsuperscript{*} \\
  Carnegie Mellon University \\
  \texttt{lynnetteng@cmu.edu} \\
  }

\begin{document}
\maketitle
\def\thefootnote{*}\footnotetext{These authors contributed equally to this work.}\def\thefootnote{\arabic{footnote}}

\begin{abstract}

Singlish, or Colloquial Singapore English, is a language formed from oral and social communication within multicultural Singapore. In this work, we work on a fundamental Natural Language Processing (NLP) task: Parts-Of-Speech (POS) tagging of Singlish sentences. For our analysis, we build a parallel Singlish dataset containing direct English translations and POS tags, with translation and POS annotation done by native Singlish speakers. Our experiments show that automatic transition- and transformer- based taggers perform with only $\sim80\%$ accuracy when evaluated against human-annotated POS labels, suggesting that there is indeed room for improvement on computation analysis of the language. We provide an exposition of challenges in Singlish annotation: its inconsistencies in form and semantics, the highly context-dependent particles of the language, its structural unique expressions, and the variation of the language on different mediums.
Our task definition, resultant labels and results reflects the challenges in analysing colloquial languages formulated from a variety of dialects, and paves the way for future studies beyond POS tagging.
\end{abstract}

\section{Introduction}
Singlish, or Colloquial Singapore English, an artifact of oral communication within a multicultural society. Understanding of the Singlish language sometimes pose challenges, because it boasts phrases and sentence structures influenced by regional languages and dialects, such as Malay and Chinese \cite{deterding2007singapore}. Thus, Singlish deviates significantly from its substrate language, English, and speakers who are only familiar with American or British variants tend to have difficulty understanding the Singlish \cite{liu2022singlish}.

Fundamental to the understanding of a language is the determination of the Parts-Of-Speech (POS) tags of words in a sentence. POS indicates the linguistic function of words in a sentence, or the role the words play in a sentence, enabling construction and understanding of sentences \cite{kupiec1992robust}. These tags are a grammatical classification that includes prepositions, noun, verb, adjectives, adverbs and so forth \cite{spacy}.  
Automatic POS tagging is an NLP research area that systematically assigns each word of a sentence to its syntactic tag \cite{chiche2022part}. These POS tags provide valuable information about grammatical structure of text, which is essential for downstream tasks such as Named Entity Recognition (NER) and machine translation. Therefore, being able to properly tag Singlish text is essential due to its eccentric grammatical structures and unclear entity boundaries. This allows us to provide better clarity of the language and the conversations between its speakers.

In this work, we express a key challenge in understanding the Singlish language: the difficulty of a POS tagging. We build a parallel Singlish dataset containing direct English translations and POS tags, with translation and POS annotation done by native Singlish speakers. We compare manually and automatically tagged POS of Singlish sentences, which reveals that current automatic taggers are still unable to understand the Singlish language. Thereafter, we provide qualitative examples and analysis to explicate the challenges of Singlish annotations, and provide a vision for opportunities of Singlish studies.

\section{Literature Review}
Singlish is an understudied language within the Natural Language Processing (NLP) field. Key NLP works in this space include the following. \citet{NUS-SMS-Corpus} constructed the NUS-SMS-Corpus which contains short messages from 23 countries, one of which is Singapore, providing a corpus of daily Singlish use.
\citet{liu2022singlish} defined a Singlish-to-English paraphrasing task as three sub-tasks: lexical level lexical level normalization, syntactic level editing, and semantic level rewriting, which achieved best performing F1 BERTScore of 87.9\%. 
\citet{wang2024seaeval} evaluated the cultural reasoning ability and cross-lingual consistency of Large Language Models using SG-Eval, a question-and-answer dataset the authors constructed that embeds cultural nuances. \citet{foo2024disentangling} clustered the Singlish-specific discourse particles to discern their contextual meaning.

Parts-Of-Speech tagging is a mature research area, with approaches ranging from using rule-based tagging \cite{brill1992simple} to statistical approaches \cite{tasharofi2007evaluation} to supervised machine learning methods for tagging \cite{lv2016corpus}. Further, \citet{toutanova2003feature} construct a dependency network to provide preceding and succeeding tag context to the word. Recently, machine learning and deep learning based POS taggers have been developed to efficiently identify words in a given sentence \cite{chiche2022part}. Hidden markov models have also been developed as an unsupervised method to perform POS tagging, especially for low-resource languages \cite{banko2004part}.  

Singlish has a unique linguistic ecology and therefore requires its own POS tags. Past work using automated tagging systems on Singlish sentences from the International Corpus of English reveals grammatical borrowings and lexical loans from other languages sources of inaccuracies as key sources of tagging inconsistencies \cite{lin2023tagging}. \citet{vyas2014pos} used the Stanford POS tagger \cite{toutanova2003feature} to tag everyday English-Hindi speech collected from the social media platform Facebook. We build on this work and perform POS tagging on Singlish texts used in everyday conversation, particularly, Short Message Services. 

\section{Dataset Construction}
\subsection{Data}
Our data is a subset of the \texttt{NUS-SMS-Corpus} \cite{NUS-SMS-Corpus}. This corpus contains 55k instances of texts from Short Messaging Service (SMS) texts. These texts are conversational in nature, and are representative of how Singlish is used for daily communications. We filter the corpus for sentences with lengths of more than 100, and then randomly sample 92 sentences for our work. 

\paragraph{Raw texts.} The 92 sentences written in Singlish obtained from the corpus, referred to as \texttt{\textbf{raw}}.

\paragraph{Translated texts.}
We provide a parallel English corpus to conduct parallel analysis. The texts are manually translated from Singlish to English by two native Singlish speakers, who have been educated in English through high school and college. The translation were cross-checked by the two annotators to preserve the semantics and intention of the raw text, with minimal edits to the raw text. This set of texts is referred to as \texttt{\textbf{translated}}.

\subsection{POS Annotation of Singlish texts}
To compare the effectiveness of automatic POS taggers on Singlish texts, we conduct both automatic and manual annotations. 

\paragraph{Manual.} A native Singlish speaker manually annotated Singlish texts with POS tags, based on the intended meaning of each word. We refer to manually annotated POS tags as \texttt{\textbf{manual}}. 

\paragraph{Automatic.} We use two pre-trained models from spaCy \cite{spacy} for automatic POS tagging. These auto-taggers are chosen as it is accessible and widely used by the community. 

\begin{itemize}
    \item \texttt{\textbf{auto-sm}}: \texttt{en\_core\_web\_sm} is a small model that employs a transition-based architecture \cite{chen-manning-2014-fast}. 
    \item \texttt{\textbf{auto-trf}}: \texttt{en\_core\_web\_trf} is deep model that is fine-tuned on the \texttt{roberta-base} architecture. 
\end{itemize}

\section{Experiments}
In this section, we conduct quantitative analysis on the performance of the automatic POS taggers against the manual tags. Then, we present qualitative examples to examine sources of errors. 

\subsection{Performance of Automatic POS taggers}

We compare the performance of automatic POS taggers with manual POS tags as labels. From table \ref{table:pos_acc}, we see that both \texttt{auto-sm} and \texttt{auto-trf} only attains an accuracy of 0.80 and 0.82. This pales in comparison to its stellar performance when tagging for English sentences, with an accuracy of 0.97 and 0.99 respectively. This comparative result suggests that there is an obvious gap in using existing English taggers for Singlish texts. 

\begin{figure*}%
    \centering
    \subfloat[\centering \textbf{auto-sm} vs \textbf{manual}]{{\includegraphics[width=0.42\linewidth]{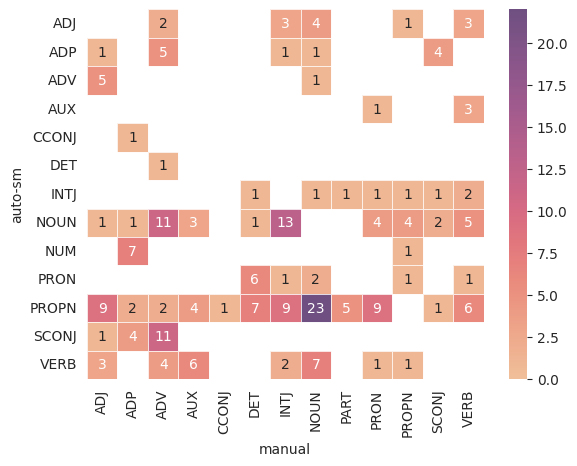} }
    \label{fig:autovmanual_manual}}%
    \qquad
    \subfloat[\centering \textbf{auto-trf} vs \textbf{manual}]{{\includegraphics[width=0.42\linewidth]{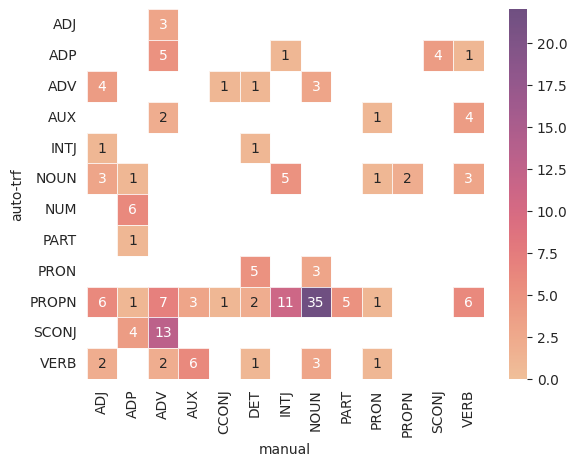} }
    \label{fig:autovmanual_trf}
    }%
    \caption{Comparison of the POS tagging by automated taggers to manual annotations.}
    \label{fig:autovmanual}%
\end{figure*}

Figure \ref{fig:autovmanual_trf} shows the discrepancies between automatic and manual tags. In general, we observe that the automatic tagger defaults to \texttt{NOUN}, and \texttt{PROPN} when the word lemma deviates from standard English. This is mostly observed in sentences containing Singlish vocabulary. For example, the Singlish sentence-final particle \textit{ba} is usually an \texttt{INTJ}, but is erroneously tagged as a \texttt{NOUN} by the \texttt{\textbf{auto-sm}}.

We obtain similar observations when using \texttt{\textbf{auto-trf}} (see Figure \ref{fig:autovmanual_trf}). \texttt{\textbf{auto-trf}} is able to tag more closely to our manual annotations, but still tend to predict \texttt{PROPN} tags when encountering Singlish vocabulary. 

\begin{table}[t]
\centering
\begin{tabular}{l|c|c}
\hline
 & \textbf{Singlish}    & \textbf{English}  \\\hline
\texttt{auto-sm}        & 0.80 & 0.97* \\ 
\texttt{auto-trf}       & 0.82 & 0.99* \\
\hline
\end{tabular}
\caption{Accuracy of POS tags for both Singlish and English data. English results (*) are referenced from official spaCy documentation \cite{spacy}.}
\label{table:pos_acc} 
\end{table} 




\subsection{Qualitative Examples}
\begin{figure*}
\centering
\includegraphics[width=\linewidth]{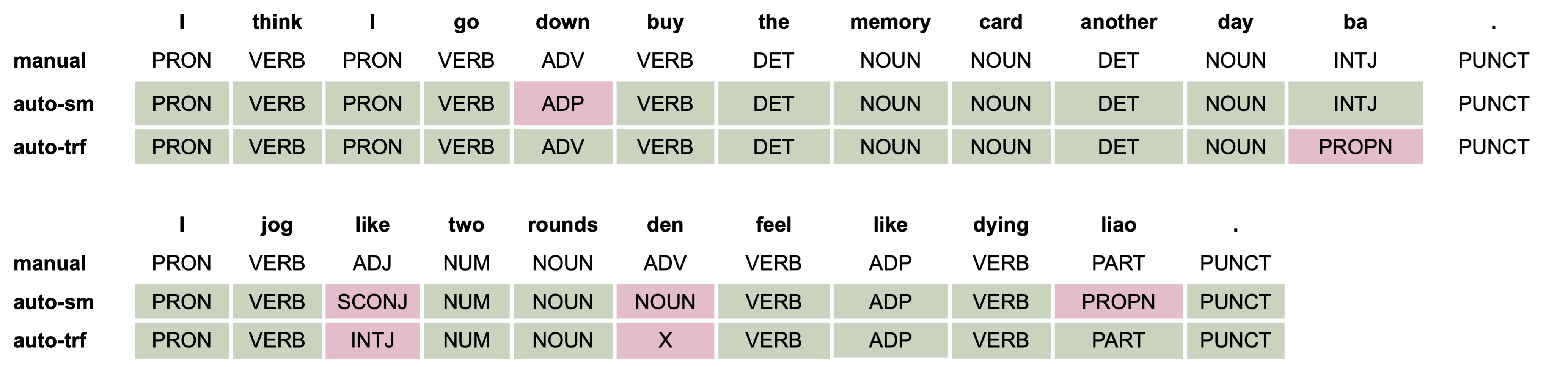}
\caption{Pink cells indicate tags that differ between manual and auto tagging}
\label{fig:eg1}
\end{figure*}

We present qualitative examples in \autoref{fig:eg1}, showcasing the difference in POS tags by the different tagging models. 

In the first example, the phrase \textit{``go down"} is a fixed expression that means ``go". This phrase borrows ``down" from Chinese \begin{CJK*}{UTF8}{gbsn}下\end{CJK*}, that translates directly to ``down", but can mean ``go" when used as a verb. For example, 

\begin{displayquote}
\begin{CJK*}{UTF8}{gbsn}我 . 下 . 餐馆 . 吃 . 面\end{CJK*} \newline
I . go . restaurant . eat . noodle
\end{displayquote}

Similarly in the phrase, ``down" does not indicate the direction of action and thus does not function as an adposition, as annotated by \texttt{\textbf{auto-sm}}. 

Next, the sentence-final particle \textit{ba} is adopted from the Chinese word, \begin{CJK*}{UTF8}{gbsn}吧\end{CJK*}. As the particle is not usually part of the English lexicon, the auto-taggers are confused of its semantic role. Note that while we refer to \textit{``ba"} as a particle, it should technically be tagged as an interjection, as it is emotive and is not syntactically related to any preceding clauses. 

In the second example, ``like"  is interpreted as ``approximately", and the authors consequentially tagged it as an adjective (\texttt{ADJ}). \textbf{\texttt{auto-sm}} instead tags ``like" as a conjunction (\texttt{SCONJ}) as in ``He is like a monkey". \textbf{\texttt{auto-trf}} tags it as an interjection \texttt{INTJ} as it interpreted it as a sentence filler. While this interpretation is reasonable, it is inaccurate in this context.  

These examples illustrate the limitations of auto-taggers due to language ambiguity and difference in vocabulary used. 

\section{Challenges in Singlish Annotation}
Singlish is an artifact of a mixture of oral and social discourse. Despite being heavily interweaved with English vocabulary, analyzing Singlish texts poses challenges to standard English NLP. In this section, we explicate three challenges in annotation: inconsistencies in form and semantics, highly context-dependent particles, structurally unique expressions. We also detail the challenge of language variation across different sub-communities. 

\subsection{Inconsistencies in Form and Semantics} \label{sec:particles}
The first challenge in annotation is the inherent inconsistencies in form and semantics of the Singlish vocabulary.

Singlish is inconsistent with English. Singlish typically condenses phrases by dropping the existential verb: \textit{``What say you?"} vs ``What do you say?; \textit{``The bus coming"} vs ``The bus is coming".

Further, the Singlish vocabulary is borrowed from a myriad of languages and dialects, which are themselves not well-documented \cite{adam2022past,wong2005you}. For example, the phrase \textit{bojio} which means ``did not invite", comes from the Hokkien phrase \begin{CJK*}{UTF8}{gbsn}无招\end{CJK*}, a phrase which does not have standardized Romanized spelling. 

Such inconsistencies confuse POS taggers trained using supervised machine learning methods, resulting in differing POS tags between automatic taggers.

The following are examples where the auto-taggers the same particle with different POS tags:
\begin{itemize}
    \item \textit{Tdy nothing to do \textbf{leh}, now playing with my laptop.} \newline \texttt{\textbf{auto-sm}} tags the particle \textit{leh} as a NOUN, while \texttt{\textbf{auto-trf}} tags as an interjection (INTJ).
    \item \textit{I reach \textbf{le} find me at e lesser ppl public phones corner.} \newline The particle \textit{le} is borrowed from the Chinese character \begin{CJK*}{UTF8}{gbsn}了\end{CJK*}, which indicates ``completion" of an event. \texttt{\textbf{auto-sm}} tags the particle \textit{le} as INTJ, while \texttt{\textbf{auto-trf}} is unable to assign a tag (X).
\end{itemize}

Further, there is also no standardised surface forms of Singlish expressions. For example, the sentence-final particles \textit{``lah"}, \textit{``lorh"}, \textit{``bah"} are synonymous with ``la", ``lor" and \textit{``ba"}. The differences in spellings commonly stem from differences in the speaker's socio-economic backgrounds.\footnote{This is an empirically observed phenomenon across the Singaporean society.} Even standard English vocabulary like ``already" can be found as \textit{``oredi"}, a pseudo-phonetic realisation of the word's Singlish pronunciation by the older generation. 

Therefore, the lack of standardisation in Singlish poses challenges in data pre-processing and model construction for computational tasks. 

\subsection{Singlish Particles are Highly Context-Dependent}

Language, by nature, is context-dependent. Singlish capitalises on context to compress multicultural expressions and convey intent. For example, \textit{``Eh join us to makan lah!"} reflects a phrase ``Hey, why not join us for a meal?" in an inviting tone. This phrase has multicultural roots: \textit{``Eh"} is influenced by the Hokkien dialect for \begin{CJK*}{UTF8}{gbsn}欸\end{CJK*}, \textit{``makan"} is a Malay word for eat, and \textit{``lah"} is a Chinese sentence-final exclamation \begin{CJK*}{UTF8}{gbsn}了\end{CJK*}.

\begin{figure}[h]
\centering
\includegraphics[width=\linewidth]{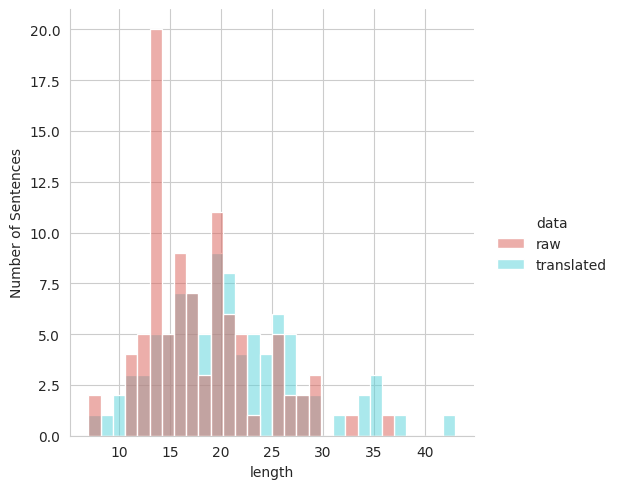}
\caption{Number of sentences vs. sentence length}
\label{fig:lengthdist}
\end{figure}

Figure \ref{fig:lengthdist} shows that for a parallel corpus, the sentence length of Singlish sentences are generally shorter than that of its English translations. Singlish express more in less words, as sentence-final particles can be used to convey many meanings, depending on its context \cite{gupta2006epistemic}. As an illustration, the meaning of the word ``one" can be different despite the same placement of the word in a sentence: \textit{I must drink coffee one} and \textit{Why you like that one}. The former is an expression of certainty; the latter an expression of slight. 

This makes the discernment of the POS tag of a Singlish particle even more difficult, as the algorithm needs to be able to identify contextual cues in order to determine its grammatical function in a sentence. In fact, Singlish particles can be so complex that there are entire articles dedicated to the analysis of a single particle, i.e. \textit{one} \cite{wong2005you}, or \textit{aiyo} \cite{wong2014culture}. 

In the \texttt{\textbf{manual}} dataset, we identified that the word ``what" has different functions in the following sentences: 

\begin{itemize}
    \item As a particle (\texttt{PART}): \textit{No nd me to intro also got lotsa admirer liao \textbf{wat}.} \newline (You don't need me to introduce you [to potential admirers] as you already have lots of admirers.)
    \item As a noun ({\texttt{NOUN}}): \textit{U muz tell me your decision so I can tell you help me bid \textbf{wat}.} \newline (You must tell me your decision so I can let you know what [courses] to help me bid for.)
\end{itemize}

In the first sentence, \textit{``what"} invokes a matter-of-fact tone, and does not directly provide additional information apart from the speaker's attitude towards the matter. Omitting the ``what" here does not fundamentally change the functional semantics of the sentence. In contrast, \textit{``what"} functions as the object in the second sentence. If omitted, ``bid" assumes the role of the object; the sentence changes its meaning to: ``You must tell me your decision so I can let you know whether or not to help me bid".

Language context is deeply tied to cultural and local norms, and thus the meaning of linguistic particles in Singlish may not necessarily be the same as that of English \cite{pennycook2006global}. NLP models need to be able to interpret the local context in order to produce accurate results. For example, the word ``banana" likely refers to a fruit in standard English, but in Singlish, the same word can refer to a Singaporean who is rather westernised. Along the same note, another Singlish phrase ``jiak kentang", which literally translates to ``eat potato" also means the same thing. 

\subsection{Singlish Expressions are Structurally Unique}
\begin{figure*}[h]
\centering
\includegraphics[width=0.85\linewidth]{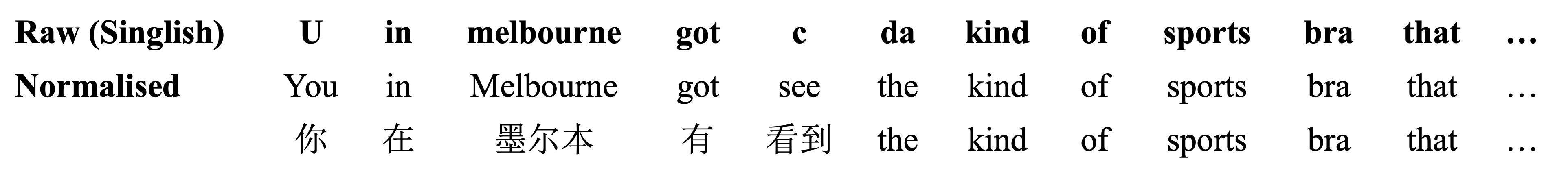}
\caption{A Singlish sentence that adopts a 1-1 translation to Chinese for part of the sentence.}
\label{fig:chinese_struct_eg}
\end{figure*}

Apart from the heavy usage of sentence-final particles, Singlish is also unique in its structure as it adopts structures from superstrate languages; one of such language is Chinese \cite{goldar2014sentence}. This results in sentences with mixed structures. Figure \ref{fig:chinese_struct_eg} illustrates an example that is partially a word-for-word translation from Chinese. With part of the English sentence being a literal translation from the Chinese language, the sentence does not follows conventional linguistic structures present in either languages, and thus cannot be interpreted with traditional linguistic methods. 

Singlish structures is an active research space in the linguistic community, as in \cite{goldar2014sentence}, \cite{leimgruber2013singapore}. A conventional Singlish sentence can adopt structures from multiple languages. However, it is challenging to apply the observed linguistics resolutions to computational analysis using heuristics or deep learning methods, due to aforementioned challenges in annotation and normalisation of data. 

\subsection{Variation in Different Sub-communities}
Within this study, the authors primarily conduct analysis on SMS texts. However, Singlish can also differ among different sub-communities. This variance is found when applied within text messages, on different social media platforms, or even when spoken orally. This difference is mainly presented in the vocabulary and expressions that are frequently used, and language used by sub-communities do not necessarily intersect. Consequentially, it is is necessary to collect and analyse texts from a wide range of sources. 

To illustrate, some phrases are exclusively used by certain communities. \textit{``Talk cock sing song"} is a phrase that means to chit-chat and gossip. The phrase is considered quite vulgar, and is mostly used by male youth when they arrange to meet for drinks. The older generation can often be heard at coffee shops saying \textit{``Limpeh ga li gong"}, which means, ``father wants to tell you" when regaling tales of their past. 

\section{Beyond POS Tagging}
Multi-, cross-lingual and low-resource NLP tasks call for universal annotations, where annotations such as POS tags, dependency relations and tree structures can be formalised within a singular system, thus ``universal". \citet{nivre-etal-2020-universal} for example, provides a cross-linguistically consistent framework for tree banking.

However, the uniqueness of Singlish expressions inadvertently results in challenges in adopting universal annotations. Apart from difficulty in identifying and tagging sentence-final particles as described in Section \ref{sec:particles}, special Singlish constructions built on standard English vocabulary pose certain difficulty in structure parsing. For example, the phrase \textit{``where got"} is commonly used as a negation or a denial of a previous statement. \cite{lauren2018analysis} observes that when ``got" appears after ``where" in a matrix clause, ``where" cannot be interpreted as a request for a location. In a sense, it becomes ambiguous how the phrase should be annotated in relation to the rest of the sentence. Canonically, ``where" is related to the head of clause with an \texttt{advmod} relation, indicating instances where the word functions as an adverb (\textit{Where do we go?}), or with an \texttt{acl:recl} relation, where it functions as a relative clause modifier (\textit{We'll see where he goes.}). 

\begin{figure*}
\centering
\includegraphics[width=\linewidth]{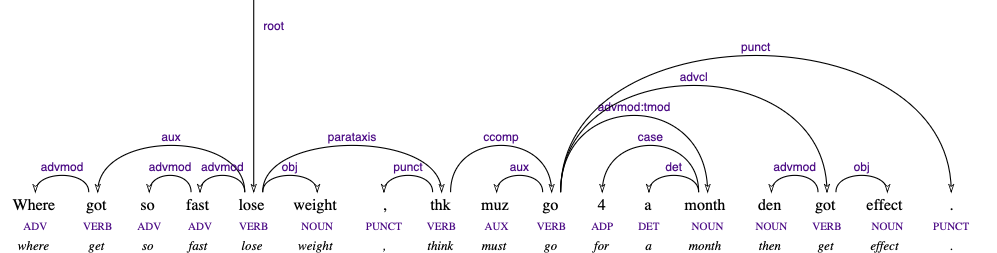}
\caption{Universal Dependency \cite{universaldependenciesUniversalDependencies} annotation for the ``where-got" construction.}
\label{fig:wheregot}
\end{figure*}

Figure \ref{fig:wheregot} shows a manually annotated dependency tree of a Singlish sentence containing the \textit{``where-got"} construction. The phrase \textit{``where got so fast lose weight"} can be interpreted in the following manner: 

\begin{itemize}
    \item It is not possible to lose weight so quickly.
    \item Can so fast lose weight meh? 
    \item Can so fast lose weight? 
    \item Where got so fast lose weight one? 
    \item Can't so fast lose weight one.
\end{itemize}

The annotators tagged \textit{``where got"} as an auxiliary to the head of the clause, as we reason that it functions more similarly to the auxiliary \textit{``can"}. However, such a decision is arbitrary at best and the annotation of Singlish constructions requires more vigorous study. Such an analytical process is involved and almost impossible to program into a automatic annotator. 

\subsection{Opportunities in Singlish Studies}
The study of Singlish opens up several avenues for the NLP community. First, the disconnect between manually and automatically produced POS tags calls for further development of manually annotated datasets with POS tags, and for specialised tagging mechanisms. 

The more accurate POS tags can be used for downstream tasks, which includes: Singlish-to-English machine translation, a task which will make the study of this language more accessible to the wider audience, including the linguistic community, the NLP community, and the general public; named-entity recognition, a task which will locate named entities and provide clear classification of these entities within the Singlish language; sentiment and stance analysis, a task which will provide insights towards the opinion of Singlish users with respect to particular events.

Finally, Singlish is an evolving language. Over time, new particles are incorporated, while some particles are phased out \cite{gonzales2023corpus}. The dynamicity of the language reflects generational changes, world and regional trends. This therefore providing opportunities for studying the social shifts from a linguistic lens.

\subsection{Dataset Availability}
There are only a few corpus curated for Singlish: the International Corpus of English - Singapore \cite{uzhICESingapore}, the SG-Eval dataset \cite{wang2024seaeval}, the CoSEM corpus \cite{gonzales2023corpus}, and the NUS-SMS-Corpus which is used in this study \cite{NUS-SMS-Corpus}. While these datasets contain up to 20,000 sentences, the lack of annotations for a variety of language tasks reduces the amount of downstream research conducted.

Therefore, to facilitate reproducibility and downstream research, we release the our dataset containing (1) manually annotated raw data, (2) translated data and (3) automatically annotated data. These data are released in \url{https://github.com/luoqichan/singlish}. 

\section{Conclusion}
English is a commonly studied language within the NLP community. However, equally as important are the studies of colloquial variants of English, such as Singlish, for greater assessbility and understanding of the writings and speech of unique groups of people. 

In our study, we construct a dataset of Singlish sentences with manually annotated POS tags. Unfortunately, current automatic POS taggers are unable to accurately tag the Singlish sentences, which may lead to cascading errors in downstream tasks.
We acknowledge that our study consists of limited annotation examples. Despite best efforts, the authors were only able to annotate and translate 92 examples. Although this is sufficient for a preliminary analysis, the dataset is limited in its usage as a computational dataset for training downstream NLP tasks. Further work is in progress in building a more comprehensive dataset.

Nonetheless, our Parts-Of-Speech annotations and investigations pave the way for further NLP research in Singlish. Further, we detail several challenges that impede its computational analysis of Singlish. This presents opportunities on potential directions of the computational study of Singlish, and encourage more native speakers to actively engage in this area of research. 

\section*{Acknowledgments}
The authors thank Russell Goh for his help with formal English translations. The authors thank the National University of Singapore Development Grant for funding the work.

\bibliography{custom}

\appendix


\end{document}